%% file: main.tex
\documentclass{article}

\usepackage{arxiv}

\usepackage[utf8]{inputenc} 
\usepackage[T1]{fontenc}    
\usepackage{hyperref}       
\usepackage{url}            
\usepackage{booktabs}       
\usepackage{amsfonts}       
\usepackage{nicefrac}       
\usepackage{microtype}      
\usepackage{lipsum}
\usepackage{enumitem}
\usepackage[numbers]{natbib}

\input{Figs/math_commands}

\usepackage{amsmath,bm,amsthm}

\theoremstyle{definition}

\usepackage{enumitem}
\usepackage{subfig}
\usepackage{xcolor,color}
\usepackage{graphicx}
\usepackage{wrapfig}
\usepackage{mathtools}
\usepackage{multirow,multicol}
\usepackage{booktabs}
\usepackage{lipsum}
\usepackage{sidecap}
\usepackage{cleveref}
\definecolor{deepmagenta}{rgb}{0.8, 0.0, 0.8}
\definecolor{applegreen}{rgb}{0.55, 0.71, 0.0}

\usepackage{letltxmacro}
\renewcommand{\eqref}{Eq.~\ref}

\definecolor{mydarkblue}{rgb}{0,0.08,0.45}
\hypersetup{
 citecolor=mydarkblue,
 linkcolor=mydarkblue,
 urlcolor=blue}
 
\usepackage[export]{adjustbox}

\usepackage[linesnumbered,ruled,vlined]{algorithm2e}

\theoremstyle{remark}

\usepackage[]{algorithm2e}

\usepackage{enumitem}

\usepackage{etoolbox}

\makeatletter
\patchcmd{\@algocf@start}
  {-1.5em}
  {0pt}
  {}{}
\makeatother

\usepackage[export]{adjustbox}

\newcommand{\cg}{\text{\rm CG}}

\newcommand{\compG}{\texttt{CompG}}
\newcommand{\dcompG}{\texttt{D-CompG}}

\newcommand{\node}{\texttt{node}}

\newcommand{\name}{\textbf{EDGE}}
\newcommand{\fea}{\text{\rm feature}}
\newcommand{\emb}{\texttt{emb}}

\usepackage{stmaryrd}

\makeatletter
\newcommand{\xMapsto}[2][]{\ext@arrow 0599{\Mapstofill@}{#1}{#2}}
\def\Mapstofill@{\arrowfill@{\Mapstochar\Relbar}\Relbar\Rightarrow}
\makeatother

\title{\Large  
Efficient Dynamic Graph Representation Learning at Scale}

\author{%
  Xinshi Chen$^{1}$, Yan Zhu$^{2}$, Haowen Xu$^{1}$, Mengyang Liu$^{1}$, Liang Xiong$^{2}$, Muhan Zhang$^{2}$, Le Song$^{1,3}$ \\
  $^{1}$Georgia Institute of Technology, $^{2}$Facebook, $^{3}$MBZUAI
}

\begin{document}

\maketitle

\begin{abstract}
    Dynamic graphs with ordered sequences of events between nodes are prevalent in real-world industrial applications such as e-commerce and social platforms. However, representation learning for dynamic graphs has posed great computational challenges due to the time and structure dependency and irregular nature of the data, preventing such models from being deployed to real-world applications. To tackle this challenge, we propose an efficient algorithm, \textbf{E}fficient \textbf{D}ynamic \textbf{G}raph l\textbf{E}arning (\name), which selectively expresses certain temporal dependency via training loss to improve the parallelism in computations. We show that \name\, can scale to dynamic graphs with millions of nodes and hundreds of millions of temporal events and achieve new state-of-the-art (SOTA) performance.
\end{abstract}

\section{Introduction}
\label{sec:intro}

\begin{wrapfigure}[25]{R}{0.33\textwidth}
\vspace{-5mm}
\includegraphics[width=0.33\textwidth]{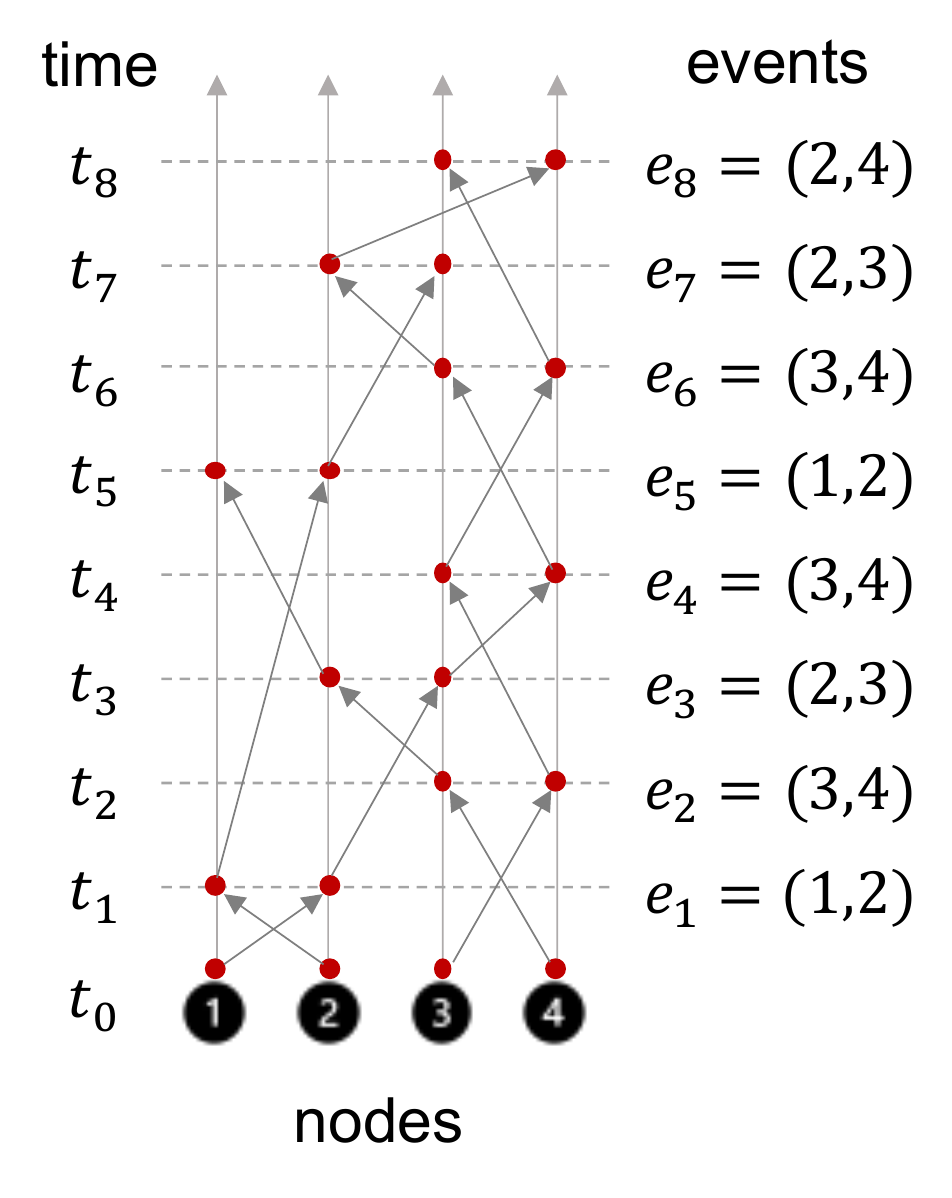}
\vspace{-7mm}
\caption{A simple example: the computational DAG of a dynamic graph model. Each red dot {\color{purple}$\bullet$}, which corresponds to a node $u\in\{1,2,3,4\}$ and a time $t\in\{t_0,\cdots,t_8\}$, indicates the computation of the temporal node embedding $\emb(u,t)$. The edges {\color{gray} $\longrightarrow$} indicate the dependencies of the computations.}
\label{fig:intro}
\end{wrapfigure}

Graph representation learning is becoming increasingly popular in addressing many graph-related applications, such as social networks, recommendation systems, knowledge graphs, biology, and so on~\citep{jacobs2001protein,otte2002social,manouselis2011recommender}. A substantial body of works has focused on \textit{static} graph neural networks (GNNs)~\citep{dai2016discriminative,hamilton2017representation,monti2017geometric,berg2017graph,ying2018graph,zhang2020revisiting}, but many real-world interactions are temporal, making \textit{dynamic} graph a more preferable model. Dynamic graphs explicitly model both the time- and node-dependency of interactions, allowing them to better represent temporal evolutions over graphs while also respecting the dynamic nature of the data.  However, unlike the active line of research on efficient algorithms for training static GNNs~\citep{hamilton2017inductive,chen2018fastgcn,chen2018stochastic,huang2018adaptive}, existing works on dynamic graph models could only be applied to small graphs.

In some sense, dynamic graphs combine the properties of sequence models and GNNs. While enjoying the representation power of both, it suffers from significantly more computational challenges. On the one hand, unlike sequence models that treat the interaction events between different nodes independently, dynamic graph models, however, introduce node dependencies, which prevents us from processing events on different nodes parallelly. On the other hand, static GNNs perform synchronized updates over the nodes, which is not allowed in the dynamic counterpart due to the time ordering constraints.

To be more specific, many dynamic graph models represent the state of a node $u\in\gV$ at time $t$ using a low-dimensional vector $\emb(u,t)\in\R^d$. To model temporal evolutions, whenever an interaction event $e=(u_1,u_2,t,r)$ occurs between nodes $u_1$ and $u_2$ at time $t$, with label $r$, the embeddings of involved nodes $u_1$ and $u_2$ will be updated by a neural operator $\gF_\theta$ in the following form:
\begin{align}
\label{eq:emb-update}
    \emb(u_1,t),\emb(u_2,t) \gets \gF_\theta\rbr{\emb(u_1, t^-), \emb(u_2,t^-), \fea(e)},
\end{align}
where the notation $\emb(u, t^-)$ represents the most recent state of the node $u$ before time $t$. Given a sequence of interaction events, \textbf{what is the computational complexity of computing all the temporal embeddings} $\emb(u,t)$\textbf{?} 

As an example, Fig.~\ref{fig:intro} shows the computational graph of the embedding updates for 8 interaction events that occurred between 4 nodes. To avoid ambiguity in terminology, we refer to a computational graph as a \textit{computational DAG} (directed acyclic graph) throughout this paper because it must be directed acyclic.

\begin{SCfigure}[][h!]
    \vspace{-2mm}
    \caption{\footnotesize Parallelization for the example in Fig~\ref{fig:intro}. JODIE takes 6 steps for the 8 events in Fig~\ref{fig:intro}. EDGE decouples 3 temporal nodes (green) from the computational graph, and treat the decoupled nodes as independent of their ancestors. Then it takes 3 steps in total to compute all temporal embeddings.}
    \includegraphics[width=0.7\textwidth]{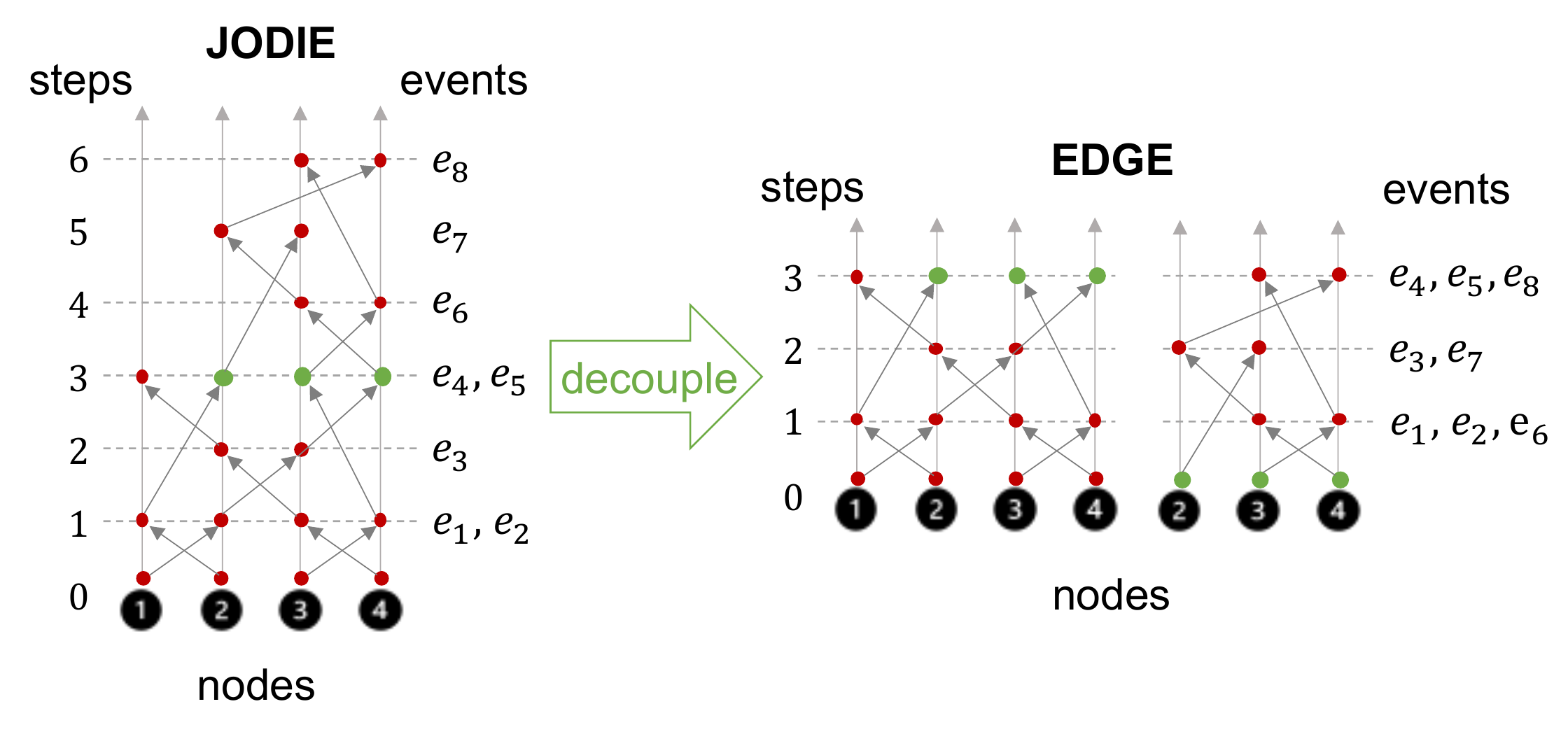}
    \label{fig:jodie-edge}
    \vspace{-6mm}
\end{SCfigure}

A naive approach can compute \eqref{eq:emb-update} sequentially according to the time ordering of interactions, which will require the \textit{number of interactions} many computational steps. To improve the computational complexity, a recent advance JODIE~\citep{kumar2019predicting} has proposed to parallelize independent computations. For example, the interaction events $e_1=(1,2)$ and $e_2=(3,4)$ in Fig.~\ref{eq:emb-update} are independent since their involved nodes do not overlap, so JODIE will perform the updates for $e_1$ and $e_2$ parallelly (please compare Fig.~\ref{fig:jodie-edge} (left) to Fig.~\ref{fig:intro}). However, the effectiveness of this strategy heavily depends on the dataset's property - to what the interaction events can be parallelized, which reveals to be far from satisfactory on real-world datasets due to the presence of many actively interacted and long-lived nodes.

To address the computational barriers that prevent the use of dynamic graph models from industrial-scale datasets, in this paper, we propose \textbf{EDGE}, which stands for \textbf{E}fficient \textbf{D}ynamic \textbf{G}raph l\textbf{E}arning.

The design of \name{} is motivated by our key finding that the computational complexity is restricted by the \textbf{longest path} in the computational DAG - no matter how we optimize the parallelization of updates within a step, it takes at least the {length of the longest path} many {sequential} steps in the end. Section~\ref{sec:algo} will go over this in greater detail. As a simple example, the computational DAG in Fig.~\ref{fig:jodie-edge} (left) has a longest path of length 6, so JODIE cannot take less than 6 steps to finish the computations. The longest path in the computational DAG represents the limit that JODIE and any other methods can reach, so EDGE is designed to carefully reduce its length 
from two aspects, as briefly summarized below.

\textit{First, EDGE selectively expresses some computational dependencies via the training loss to achieve better parallelism in computation.} To be more specific, EDGE decouples some nodes (called \textbf{d-nodes}) from the computational DAG to remove certain computational dependencies; and these dependencies will be added back via the training loss.
For instance, the d-nodes (green-colored  {\color{applegreen}$\bullet$}) in Fig.~\ref{fig:jodie-edge} are divided into two nodes each, which separates a d-node's descendants from its ancestors, shortening the longest path in the computational DAG. However, a naive decoupling strategy could lead to a sub-optimal model due to the ignored dependencies. In Section~\ref{sec:algo}, we will explain how we carefully design the decoupling strategy by answering the following two questions, which are critical to its success.
\begin{itemize}[wide,nolistsep]
    \item How to select the most effective set of d-nodes for shortening the longest path?
    \item How to modify the training loss to compensate for the removed computational dependencies caused by the d-nodes?
\end{itemize}

\begin{wrapfigure}[11]{R}{0.28\textwidth}
\vspace{-5mm}
\includegraphics[width=0.28\textwidth]{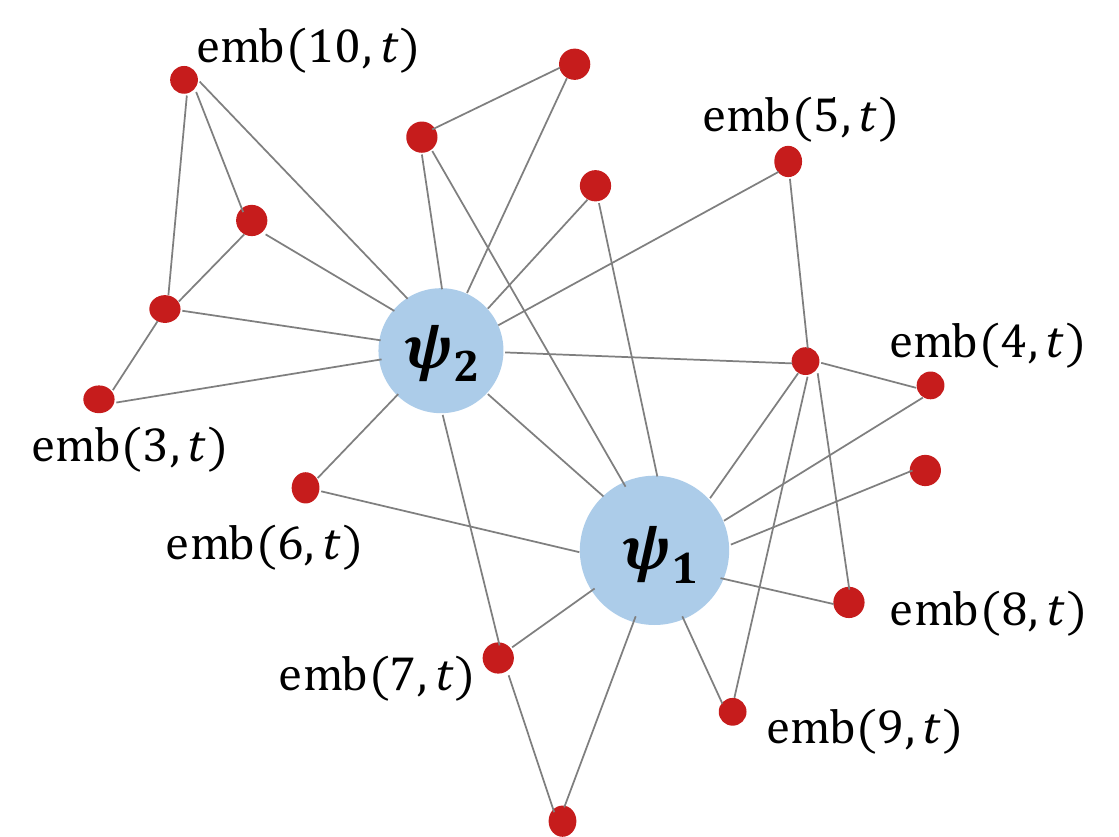}
\vspace{-4mm}
\caption{\small Scale-free network.}
\label{fig:sf}
\end{wrapfigure}
\textit{Second, EDGE assumes the \textbf{convergent states} of actively interacted nodes and models them by static embeddings.} As visualized in Fig.~\ref{fig:sf}, many realistic datasets contain  scale-free interaction networks~\citep{leskovec2005graphs}, where most nodes have a small number of interactions but a few high-degree nodes are actively interacted. Examples include celebrities on Twitter who have lots of followers, and popular items on Amazon which are frequently clicked. While these active nodes contribute a large amount of edges to the graph, their embedding states tend to be static. The reasoning behind is mainly two-fold.  First, an active node's interactions with other nodes give us \textit{diminishing} returns in terms of their information value in understanding this node's characteristics.  Second, from the optimization perspective, a recurrent operator such as $\gF_\theta$ in \eqref{eq:emb-update} often has \textit{smaller derivatives} in deeper layers, which is also the cause of the gradient vanishing problem~\citep{pascanu2013difficulty}. With the operator applied repeatedly, embeddings of active nodes will eventually evolve slowly. With this motivation, EDGE represents active nodes by node-specific \textit{static embeddings} $\vpsi_u$ during training, while keeping other nodes dynamic. The effect from other nodes to active nodes are implicitly incorporated through the gradient updates during training. After being trained, the operator $\gF_\theta$ could still be applied to active nodes during testing phase. This static treatment of active nodes proves to be very effective on realistic datasets.

In summary, these two designs jointly reduce the path length in the training computational graph, making EDGE scalable to realistic datasets. 
We conduct comprehensive experiments on several user-item interaction datasets, including two large-scale datasets, to compare EDGE to a wide range of baselines, showing the SOTA performances of EDGE in both accuracy and prediction efficiency.

\textbf{Related work.} Although learning on dynamic graphs is relatively recent~\citep{rahman2018dylink2vec, goyal2020dyngraph2vec,sankar2020dysat,rossi2020temporal,xu2020inductive}, various models were actively proposed to represent temporal patterns via the time ordering of interaction events \citep{ibrahim2015link,dai2016deep,kumar2019predicting} or from temporal aligned graph snapshots \citep{goyal2020dyngraph2vec, sankar2020dysat}. A large subset of such models that updates node-wise embeddings through a neural network when new
interactions appear~\citep{kumar2019predicting,ibrahim2015link,dai2016deep,trivedi2017know,ma2020streaming} is most relevant to this paper. They vary from each other mainly by designing different architectures and incorporating different information. The method proposed in this paper could be generally applied to improve the training efficiency of these models. For a more comprehensive summary of other dynamic models, we refer the audience to \citep{rossi2020temporal} which summarizes different models via a general framework.

\section{Model}

In this paper, we focus on modeling the interactions between two groups of nodes, $\gV_1$ and $\gV_2$, but the technique could easily be applied to other scenarios involving fewer or more nodes in each temporal event.

We denote an interaction event between nodes $u_1\in\gV_1$ and $u_2\in\gV_2$ by $e=(u_1,u_2,t,r)$, where $t$ is the time and $r$ is a label, which can indicate click/non-click in user-item interaction networks, for example. 
In this section, we will illustrate how a dynamic graph model is used to describe the evolution of node embeddings over time when a sequence of interaction events are observed.

\subsection{Embedding Evolution}

In many dynamic graph models, two crucial components for modeling the evolution of node embeddings are 
\begin{enumerate}[nolistsep]
    \item the initial embeddings $\emb(u,0)$ of the nodes; and
    \item the neural operator $\gF_\theta$ for updating them (via \eqref{eq:emb-update}).
\end{enumerate}
In EDGE, a key consideration is the convergent states of active nodes. Therefore, it brings in a new component:
\begin{enumerate}[nolistsep]
    \item[3.] \textbf{the node-specific static embeddings for active nodes}.
\end{enumerate}
To be more specific, Given a degree threshold $n_1^*$, we say a node $u_1$ is active if it has more than $n_1^*$ interactions in the training set. We can denote the set of active nodes by $\gV_1^{act}\subseteq \gV_1 $ and define $\gV_2^{act}\subseteq \gV_2$ in a similary way based on a threshold $n_2^*$. In EDGE, these active nodes' embeddings are modeled by static embeddings
\begin{align}
    \emb(u,t) = \vpsi_u,\quad \forall t\leq T_{train},\quad \forall u\in \gV_1^{act} \cup \gV_2^{act},
    \label{eq:act}
\end{align}
where each $\vpsi_u$ is a learnable vector, and $T_{train}$ is the final time-stamp in the training dataset. The other `inactive' nodes are evolved by the neural operator $\gF_\theta$ when the interactions $e=(u_1,u_2,t,r)$ occur as follows.
\begin{align}
    &\forall t\leq T_{train},\quad \forall u_1\in \gV_1\setminus \gV_1^{act} \nonumber \\
    &\emb(u_1,0) = \vxi_1\\
    &\emb(u_1,t)\gets \gF_{\theta_1}\rbr{\emb(u_1, t^-), \emb(u_2,t^-), \fea(e)} \label{eq:inact2}\\
    &\forall t\leq T_{train},\quad \forall u_2\in \gV_2\setminus \gV_2^{act}  \nonumber\\
    &\emb(u_2,0) = \vxi_2\\
    &\emb(u_2,t)\gets \gF_{\theta_2}\rbr{\emb(u_2, t^-), \emb(u_1,t^-), \fea(e)}, \label{eq:inact}
\end{align}
where $\vxi_1,\vxi_2$ are learnable vectors that define the common initializations for inactive nodes in $\gV_1$ and $\gV_2$, and we use different parameters $\theta_1$ and $\theta_2$ in the operator $\gF$ that updates these two groups of nodes. The architecture of $\gF$ is not the main focus of this paper, so we simply adopt a gated recurrent unit (GRU) \citep{chung2014empirical} based architecture.  The notion $\fea(e)$ can be any available event features. Examples include contents of posts on Twitter, static features of items on shopping platforms, etc. Lastly, the notation $t^-$ in $(u,t^-)$ refers to the time-stamp of the last event occured to $u$ before $t$. 

\eqref{eq:act} to \eqref{eq:inact} jointly define the overall embedding evolution model in EDGE. A few remarks and important discussions for this model are presented in the following.

\textbf{Efficiency gain.} The reduction of computational dependencies achieved by representing active nodes using  static embeddings is significant, because the active nodes contribute a substantial number of edges to the network. The only computational sacrifice here is the memory cost for introducing the additional set of learnable vectors $\{\vpsi_u\}$ for the active nodes. However, in our experiments, this does not cause any problems. Every transductive graph neural network, by comparison, will require a memory cost greater than that.

\textbf{Accuracy.} The static treatment of active nodes does damage the model's performance, which will be verified by extensive experiments in Section~\ref{sec:experiment}. This static strategy is well-thought-out and supported by a number of arguments below.
\begin{enumerate}[wide,nolistsep]
    \item If we view the embedding evolution as a refresh of our understanding about the features of a node, the information value of a new interaction for an active node is minimal.
    \item Optimization is as important as the model. 
    With the reduced computational dependencies, the training optimization becomes easier, which might also be the reason why we often observe an improvement in model performance in our experiments.
    \item From the view of variance-bias trade-off, it is reasonable to use a transductive (i.e., node-specific) embedding for active nodes and use an inductive model for inactive nodes.
    \item During the test phase, we could still apply the learned operator $\gF_\theta$ to evolve the embeddings of active nodes according to their newly observed events.
\end{enumerate}

\subsection{Prediction Model and Loss Function}

After computing the node embeddings, a predictive model extras useful features to archieve the desired prediction. Our experiments are mainly about predicting the next interacted node in $\gV_2$ for the nodes in $\gV_1$, so we employ a neural layer $\gH_\theta$ to predict the embedding of the next interacted node.  Here we abuse the notation $\theta$ to generally represent parameters in neural networks. Inspired by the design in \citep{kumar2019predicting}, we employ a prediction layer in the following form:
\begin{align*}
    \vh_{u_1,t}\gets\gH_\theta(\emb(u_1,t), \fea(u_1), \fea(\node(u_1,t^-))),
\end{align*}
where $\node(u_1,t^-)\in\gV_2$ is the most recently interacted node of $u_1$ before time $t$. Here $\fea(\cdot)$ denotes any available node features. In our experiments, $\gH_\theta$ is simply a multi-layer perceptron (MLP).

With the prediction embedding $\vh_{u_1,t}$, one can then measure its similarity to embeddings of nodes in $\gV_2$. 
In our experiments, we measure its similarity $s_t(u_1,u_2)$ to a node $u_2$ using inner product, and use cross-entropy loss (equivalently, the softmax loss) for each event $e$ in the training set $\gD_{train}$:
\begin{align}
    \gL(e=(u_1,u_2,t,r)\,;\Theta) = - \log\rbr{
    \frac{\exp\rbr{s_t(u_1,u_2)}}{\sum_{u_2'\in\gO(u_1,t)} \exp\rbr{s_t(u_1,u_2')}}
    }
\end{align}
where $\Theta:=\{\theta_1,\theta_2,\vxi_1,\vxi_2,\{\vpsi_u:u\in\gV_1^{act}\cup\gV_2^{act} \}\}$ includes all parameters. $\gO(u_1,t)$ is a set of `negative' nodes that $u_1$ did not interact with at $t$. In our experiments, this set is randomly sampled from $\gV_2$.

\section{Efficient Training Algorithm}
\label{sec:algo}

The bottleneck of previous training methods is the under-utilization of GPUs since the computation is hard to parallelize. We find that the computational complexity is limited by the longest path in the computational DAG, which motivates our design of \textit{d-nodes} for shortening it. We will describe the details in this section.

\subsection{Why Longest Path?}
Why is the computational complexity limited by the longest path in the computational DAG? To answer this question, we would like to refer the audience to the literature of parallel algorithm analysis. Briefly speaking,
\begin{itemize}[leftmargin=5mm,nolistsep]
    \item The so-called \textit{work-depth model} of parallel computation was originally introduced by \cite{shiloach1982n2log} in 1982, which has been used for many years to describe the performance of parallel algorithms.
    \item \textit{Depth} is defined as the longest chain of sequential dependencies in the computation~\cite{blelloch1996programming}.
    \item Minimizing the depth is important in designing parallel algorithms, because the depth determines the \textit{shortest possible execution time} \cite{mccool2012structured}.
\end{itemize}
When the computations of a dynamic graph are viewed as the operations in a parallel algorithm, then the ``longest path in the computational DAG'' corresponds to the ``depth'' in the work-depth model, which determines the shortest possible execution time.

\subsection{A Simplified Illustration of D-nodes}
\label{sec:simp-example}

Before going into the details of the \textbf{d-nodes}, we would like to first explain the high-level idea through a highly simplified example. Instead of the very complex dynamic graph model, consider a 6-layer feed-forward network $\sigma(w_6\,\sigma(w_5\,\sigma(w_4\,\sigma(w_3\,\sigma(w_2\,\sigma(w_1 x))))) )$. \\
Clearly, its computational DAG is a path
\begin{align*}
    x\rightarrow h_1\rightarrow h_2\rightarrow h_3\rightarrow h_4 \rightarrow h_5\rightarrow h_6
\end{align*}
where $h_i=\sigma(w_i h_{i-1})$ is the $i$-th hidden layer output. Clearly, it requires 6 sequential steps to finish the computations layer by layer. However, we can speed it up to 3 sequential steps by decoupling this path through the node $h_3$. 

More precisely, we \textbf{select $h_3$ as the d-node and add a learnable vector $\phi_3$ for it}. Then we can perform the computations $h_3=\sigma(w_3\,\sigma(w_2\,\sigma(w_1 x)))$ and $h_6=\sigma(w_6\,\sigma(w_5\,\sigma(w_4 \phi_3)))$ {parallelly}. The new computational DAG becomes 2 independent paths:
\begin{align*}
    x\rightarrow h_1\rightarrow h_2\rightarrow h_3\\
    \phi_3\rightarrow h_4 \rightarrow h_5\rightarrow h_6
\end{align*}
and their lengths are 3. The key is that $\phi_3$ is a learnable vector that does not depend on any other nodes, so we can start to use $\phi_3$ for computing $h_4,h_5,h_6$ before we obtain $h_3$.

The final problem is, with this decoupling, it seems the model becomes different from the original 6-layer feed-forward network. Fortunately, to maintain the consistency, we only need to enforce the constraint $h_3=\phi_3$ during the optimization. Apparently, if $h_3=\phi_3$ is satisfied, the models before and after the decoupling are equivalent. Therefore, we can perform gradient updates to optimize both the network parameters $\vw=(w_1,\cdots,w_6)$ and $\phi_3$ by solving the following constrained optimization
\begin{align}
    \min_{\vw,\phi_3} \text{loss}(h_6) \quad \text{subject to}\quad  h_3=\phi_3.
\end{align}
Experimentally, we enforce the constraint softly by:
\begin{align}
    \min_{\vw,\phi_3} \text{loss}(h_6)+\frac{\alpha}{2}  \|h_3-\phi_3\|_2^2.
\end{align}

To summarize, the key idea of d-nodes is to \textbf{selectively express some dependencies in computation by dependencies in optimization (via training loss)}. 

In this example, the computational DAG simply consists of paths. In the case of a dynamic graph, the computational DAG is far more complex. However, the algorithm follows the same logic. We will present the details in the following sub-sections.

\subsection{Step 1: Selection of D-nodes}

In the example in Section~\ref{sec:simp-example}, the computational DAG is a path, so selecting the mid-point $h_3$ as the d-node is very effective in shortening the path. However, in the case of dynamic graph models, the computational DAG is a lot more complex due to the node dependency and time dependency (e.g., Fig.~\ref{fig:intro}).

Ideally, we should choose a small number of d-nodes that can effectively shorten the longest path in the computational DAG. Therefore, in this section, we explore the answer of:

\begin{quote}
    \textit{Given a limited quota $K$, how to choose the most effective collection of $K$ d-nodes?}
\end{quote}

To begin with, we construct the computational DAG for computing all temporal node embeddings $\emb(u,t)$ through the outlined steps in Algorithm~\ref{algo:comp-g} and denote it by $\compG=(\gV_\cg,\gE_\cg)$. The goal is to decouple a subset of $K$ nodes $\gV_\cg^D\subseteq\gV_\cg$ from the computational DAG to shorten the longest path.
\begin{algorithm}
$\gV_\cg = \gV_1^{act}  \cup \gV_2^{act} \cup \{(u,0):u\in \gV_1\setminus\gV_1^{act} \cup  \gV_2\setminus\gV_2^{act}\}$\; 
$\gE_\cg = \emptyset$\;
 \For{$e=(u_1,u_2,t,r)\in\gD_{train}$}{
    Execute for both $i=1$ and $i=2$:\\
    \If{$u_i\in\gV_i\setminus\gV_i^{act}$}{
    Add the node $(u_i,t)$ to $\gV_\cg$\;
    Add the edge $\rbr{(u_i,t^-), (u_i,t)}$ to $\gE_\cg$\;
    $j=3-i$\;
    \uIf{$u_j\in \gV_j\setminus\gV_j^{act}$}{
    Add the edge $\rbr{(u_j,t^-),(u_i,t)}$ to $\gE_\cg$\;
    }
    \Else{Add the edge $\rbr{u_j,(u_i,t)}$ to $\gE_\cg$\;}
    }
 }
 \KwOut{$\compG=(\gV_\cg,\gE_\cg)$\;}
 \caption{Build computational DAG}
 \label{algo:comp-g}
\end{algorithm}

\begin{algorithm}
    \KwIn{$\gV_\cg,\gE_\cg$, $K$\;}
     Initialize $\gV_\cg^D=\emptyset$\;
     \While{$k < K$, $k$++}{
      Find a longest path $p$ in the DAG $(\gV_\cg,\gE_\cg)$\;
      Find the center node $i$ in the path $p$\;
      Add $i$ to $\gV_\cg^D$\;
      Add a new node $i'$ to $\gV_\cg$\;
      \For{$j \in \{j | (i, j) \in \gE_\cg \}$ }
      {Remove edge $(i, j)$ from $\gE_\cg$\;
      Add edge $(i', j)$ to $\gE_\cg$\;
      }
     }
      \KwOut{$\gV_\cg^D$\;}
      \KwOut{New graph $\dcompG=(\gV_\cg,\gE_\cg)$\;}
     \caption{Select $K$ d-nodes from $\compG$}
     \label{algo:dnodes}
\end{algorithm}

Mathematically, given the computational DAG, $\compG=(\gV_\cg,\gE_\cg)$, the d-node selection problem can be formulated as the following minmax integer programming (IP):
\begin{gather*}
    \min_{\{d_i\in\{0,1\},\ell_i\in\sN\}_{i\in\gV_\cg} } \Big\{ \max_{i\in\gV_\cg}\ell_i\Big\}\quad
    \\ \text{s.t. }
    \begin{cases}
    \ell_i \geq (1-d_j)\ell_j+ 1& \forall (j,i) \in \gE_\cg \\
    \sum_{i\in\gV_\cg} d_i \leq K
    \end{cases},
\end{gather*}

where the binary variable $d_i$ indicates whether the node $i\in\gV_\cg$ is selected as a d-node, and $\ell_i$ measures the longest distance between node $i$ and root nodes in the decoupled graph.

Given the extremely large problem size of this IP (the number of nodes in $\gV_\cg$ is proportional to the number of events in the dataset), we cannot use existing packages such as Gurobi~\citep{gurobi2018gurobi} to solve it.  Therefore, we design a greedy heuristic to give an approximate solution, which reveals to be effective in experiments. 

The algorithm steps are outlined in Algorithm~\ref{algo:dnodes}. Briefly speaking, it repeatedly finds the middle-point of the longest path in the computational DAG, takes it as a d-node, and then adjusts edges from this node, until $K$ d-nodes are found.  The idea of this algorithm is similar to alternative minimization maximization, but solving a sub-problem each time sequentially for $k=1,\cdots,K$:

\begin{enumerate}[leftmargin=0.5cm]
    \item[(i)] \textit{Optimize with fixed $\{d_i\}$}: Suppose values of $\{d_i\}$ are given by current selections $\{i^1,\cdots,i^{k-1}\}$. Then the node that optimizes the inter maximization $i^*=\argmax_{i\in\gV_\cg}\ell_i$  is apparently in the longest path $p$ in Algorithm~\ref{algo:dnodes}. Furthermore, for all edges in this path, the corresponding constraints in the IP  are binding (i.e., equality $\ell_i=(1-d_j)\ell_j+1$ holds for all $(j,i ) \in p $), and removing all other constraints does not change the optimal value $\ell_{i^*}$. Therefore, we select the next d-node by solving a sub-problem identified by this longest path $p$.
    \item[(ii)] \textit{Select $i^k$ on sub-problem}: This step selects a d-node $i^k$ and makes $d_{i^k}=1$, which is accomplished by solving the IP with the subset of constraints identified by edges in the longest path $p$. As pointed out in Algorithm~\ref{algo:dnodes}, the $i^k$ that optimizes the sub-problem is the center node in the path $p$.
\end{enumerate}

It is notable that the longest path of a DAG can be found in linear time, and Algorithm~\ref{algo:dnodes} only needs to run once as a pre-processing step.

\subsection{Step 2: Express Dependency by Constraints}
\label{sec:constraint}

From now on, we index each selected d-node by $(u,t)\in\gV_\cg^D$ since it has a unique correspondence to a temporal state $\emb(u,t)$.

Each d-node $(u,t)$ plays a role of detaching the events happened to $u$ after $t$ from those before $t$. Similar to adding the vector $\phi_3$ for $h_3$ in Section~\ref{sec:simp-example}, the decoupling is realized by creating an \textit{additional embedding} $\vphi_{u,t}$ for each d-node, which will be used as a replacement of $\emb(u,t)$ for any events after $t$ that involve the state of $(u,t)$. An apparent benefit is $\vphi_{u,t}$ can be used for computing future evolutions {before} obtaining the embedding $\emb(u,t)$. 

It is important that the decoupling of the d-nodes should not break the original dependencies in the model. Fortunately, it is easy to observe that involving the d-nodes will not change the obtained node embeddings if its associated two embeddings are equal:
\begin{align*}
    \vphi_{u,t} = \emb(u,t),\quad \forall (u,t)\in\gV_\cg^{D}.
\end{align*}

Therefore, during the training phase, EDGE solves a constrained optimization:
\begin{align*}
   \min_{\{\vphi_{u,t}\},\Theta} \frac{1}{|\gD_{train}|}\sum_{e\in\gD_{train}}\gL(e\,;\Theta,\{\vphi_{u,t}\}) \\ \quad \text{subject to}\quad \vphi_{u,t} = \emb(u,t),\quad \forall (u,t)\in\gV_\cg^{D}.
\end{align*}
Various algorithms are available for solving such a constrained optimization problem. For example, one can use Lagrangian method with quadratic penalty, and solve it by alternative primal-dual updates. In our experiments, we enforce the constraints softly, by adding a quadratic penalty $\frac{\alpha}{2} \|\vphi_{u,t} - \emb(u,t)\|_2^2$ to the loss whenever the node $\emb(u,t)$ is reached in the computational DAG.

\section{Experiments}
\label{sec:experiment}

We conduct experiments on four public datasets. Two of them are large datasets close to industrial scale, on which we show the scalability and prediction accuracy of EDGE,  by comparing to a diverse set of SOTA methods. The other two are comparatively smaller datasets, on which all dynamic graph baselines are runnable, so they are used for comparison to them. Implementation of EDGE will be publicly available upon acceptance.

\textbf{Datasets} include Taobao~ \citep{pi2019practice}, MovieLens-25M (ML-25M) \citep{harper2015movielens}, lastfm-1K~\citep{Celma:Springer2010}, and Reddit~\citep{baumgartner2020pushshift}. Dataset statistics are summarized in Table~\ref{tb:stats}. Taobao and ML-25M are large-scale. Especially, Taobao contains about 100 million interactions.
All datasets provide timed user-item interaction data where for each interaction, user ID, item ID, timestamp, item feature, and event feature (if available) are given. For Reddit, subreddits are considered as items. For ML-25M, we ignore the ratings and simply use it as interaction data. For Taobao and ML-25M, we sort the interactions by timestamp, and use a (0.7,0.1,0.2) data-split for training, validation, and testing, respectively. For Lastfm-1K and Reddit, we use the filtered version given by~\citep{kumar2019predicting} and follow their data-split of (0.8,0.1,0.1). More details and downloadable links to the datasets are provided in Appendix~\ref{app:dataset}.

\begin{table}[h!]
\centering
\caption{Dataset Statistics}
\label{tb:stats}
    \centering
    \begin{tabular}{|@{\hspace{1mm}}c@{\hspace{2mm}}c@{\hspace{2mm}}c@{}c@{\hspace{1mm}}|}
    \hline
        Dataset & \#users & \#items & \#interactions \\
    \hline
        lastfm-1K & 1,000 & 980 & 1,293,103 \\ Reddit & 10,000 & 984 & 672,447 \\
        ML-25M & 162,538 & 59,048 & 24,999,849 \\
        Taobao & 987,975 & 4,111,798 & 96,678,667 \\
    \hline
\end{tabular}
\end{table}

\textbf{Baselines.} We compare EDGE to a wide range of baselines spanning 3 categories: \textit{(i) Dynamic graph models} include JODIE~\citep{kumar2019predicting}, dynAERNN~\citep{goyal2020dyngraph2vec}, TGAT~\citep{xu2020inductive}, CTDNE~\citep{nguyen2018continuous}, and DeepCoevolve~\citep{dai2016deep}. The first three were proposed recently and more advanced. \textit{(ii) GNNs}: GCMC~\citep{berg2017graph} is a SOTA graph-based architecture for link prediction in user-item interaction network. GCMC-SAGE is its stochastic variant which is more efficient, using techniques in GraphSAGE~\citep{xu2020inductive}. GCMC-GAT is another variant based on graph attention networks~\citep{velivckovic2017graph}. \textit{(iii) Deep sequence models}: SumPooling is a simple yet effective model widely used in industry. GRU4REC~\citep{hidasi2015session} is a representative of RNN-based models. SASRec~\citep{kang2018self} is a 2-layer transformer decoder-like model. DIEN~\citep{zhou2019deep} is an advanced attention-based sequence model. MIMN~\citep{pi2019practice} is a memory-enhanced RNN for better capturing long sequential user behaviors. It is a strong baseline and the Taobao dataset were publicized by the authors.  

\begin{table}[ht]
    \centering
    \caption{Overall performance. EDGE uses 160 and 20 d-nodes per batch on Taobao and ML-25M respectively. ${}^*$~indicates that the released implementation is improved by us to make the method either runnable on these large datasets or better adapted to the evaluation metric, so the results are expected to be better than the original version. Modifications are described in Appendix~\ref{sec:baseline-mod}. `-oom-' stands for out of memory.}
    \label{tab:results}
    
    {\begin{tabular}{|@{\hspace{1mm}}c@{}c|c|c@{\hspace{2mm}}c|c@{\hspace{2mm}}c@{\hspace{1mm}}|}
    \hline
    & & \multirow{2}{*}{method} & \multicolumn{2}{c|}{Taobao} & \multicolumn{2}{c@{\hspace{1mm}}|}{ML-25M}
    \\
      & &  & MRR  & Rec@10   & MRR  & Rec@10\\
    \hline \hline
    \multicolumn{2}{|c|}{\multirow{3}{*}{\rotatebox{90}{\it GNNs }}} & GCMC$^* $ & 0.303 & 0.542 & 0.171 & 0.378 \\
    \multicolumn{2}{|c|}{}                      & GCMC-SAGE$^* $ & 0.149 & 0.366 & 0.109 & 0.264 \\
    \multicolumn{2}{|c|}{}                      & GCMC-GAT$^* $ & 0.230 & 0.494 & 0.185 & 0.404  \\
    \hline \hline
    \multirow{6}{*}{ \rotatebox{90}{\it\,Sequence}}& \multirow{6}{*}{ \rotatebox{90}{\it \,Models }} & SumPooling & 0.415 & 0.664 & 0.302 & 0.591\\
      & & GRU4REC$^* $ & 0.546  & 0.777  &{\color{blue}0.364}  & {\color{blue}0.657}  \\
      & & DIEN$^* $ & 0.605  & {\color{blue}0.834} & 0.356  & 0.638  \\
      & & MIMN$^* $ & {\color{blue}0.607} & 0.828  & 0.363  & 0.653   \\
      & & SASRec$^* $ & 0.488 & 0.702 & 0.360 & 0.653\\
    \hline \hline
       \multirow{3}{*}{\rotatebox{90}{\it Dynamic }} & \multirow{3}{*}{\rotatebox{90}{\it Graphs \,}} &  dynAERNN  & -oom- & -oom- & 0.249 & 0.509 \\
             & & TGAT & 0.016 & 0.025 &0.114 & 0.219 \\
      & &  JODIE$^* $ & 0.454  & 0.680  & 0.354  & 0.634  \\ 
      & & \name & {\color{red}0.675} & {\color{red}0.841} & {\color{red}0.397} & {\color{red}0.673}\\
        \hline
    \end{tabular}}
\end{table}

\textbf{Configuration.} In all experiments, both node embedding dimension and feature embedding dimension are 64. Only Reddit contains event features. Other datasets may provide categorical item features. We experimentally find that including item ID as an additional item feature is generally helpful. 
To conduct stochastic optimization, we sort the events by time and divide them into 300 batches, so that each batch contains a sub-graph within a continuous time window. To use d-nodes, we choose $K$ d-nodes per batch. The threshold $n^*$ for active nodes is chosen to be 200 for users and 100 for items on both Taobao and ML-25M. The other two datasets are filtered and small, so threshold is not necessary. We use $n^*=\infty$ for both datasets (which means no static embedding is used for any nodes). Besides, we also tried using $n^*=400$ for Reddit and $n^*=500$ for lastfm-1K.

\textbf{Evaluation metric.} We measure the performance of different models in terms of the mean reciprocal rank (MRR), which is the average of the reciprocal rank,
and recall@10. For both metrics, higher values are better. On Taobao and ML-25M, the ranking of the truly interacted item in each event is calculated
among 500 items where the other 499 negative items are uniformly sampled from the set of all items. On Reddit and lastfm-1K, ranks are over all items. In the tables in this section, we mark the best performing method  by {\color{red}red} color and the second best by {\color{blue}blue} color. 

\begin{figure}[b]
    \centering
    \includegraphics[width=0.4\textwidth]{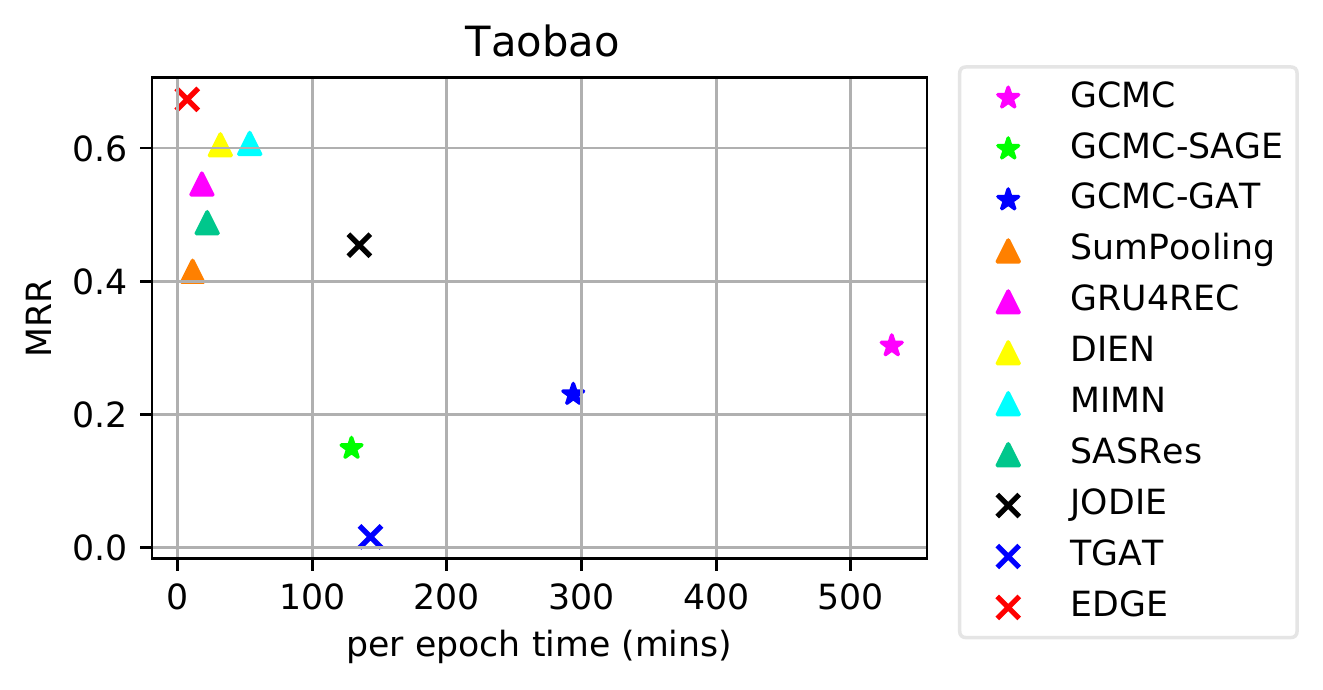}~
    \includegraphics[width=0.4\textwidth]{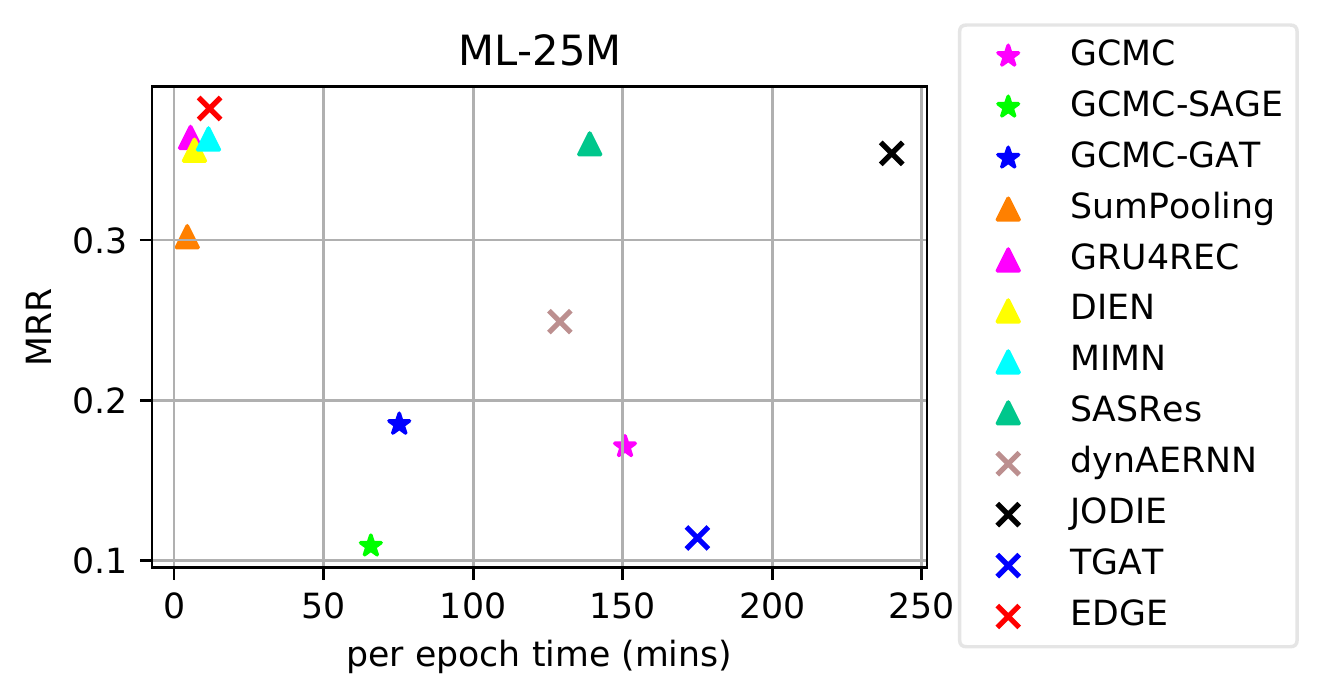}
    \caption{ Training time per epoch and MRR performance on test set. This figure shows the comparison in both efficiency and accuracy.}
    \label{fig:time-mrr}
\end{figure}

\textbf{Result 1: On large datasets.}

\textit{(i) MRR and Rec@10}: The overall performances on Taobao and ML-25M are summarized in Table~\ref{tab:results}. EDGE has consistent improvements compared to all baselines. Improvements of 11.2\% and 9.1\% over the second best method are achieved in terms of MRR. In fact, the released implementations of many baselines cannot run or give reasonable performances on these large datasets. As indicated by symbols * in Table~\ref{tab:results}, we improve the baselines to allow them to have more advantages. Please refer to Appendix~\ref{sec:baseline-mod} for details.

\begin{figure*}[t]
    \centering
    \begin{tabular}{@{}c@{}c@{}c@{}}
    \includegraphics[width=0.3\textwidth]{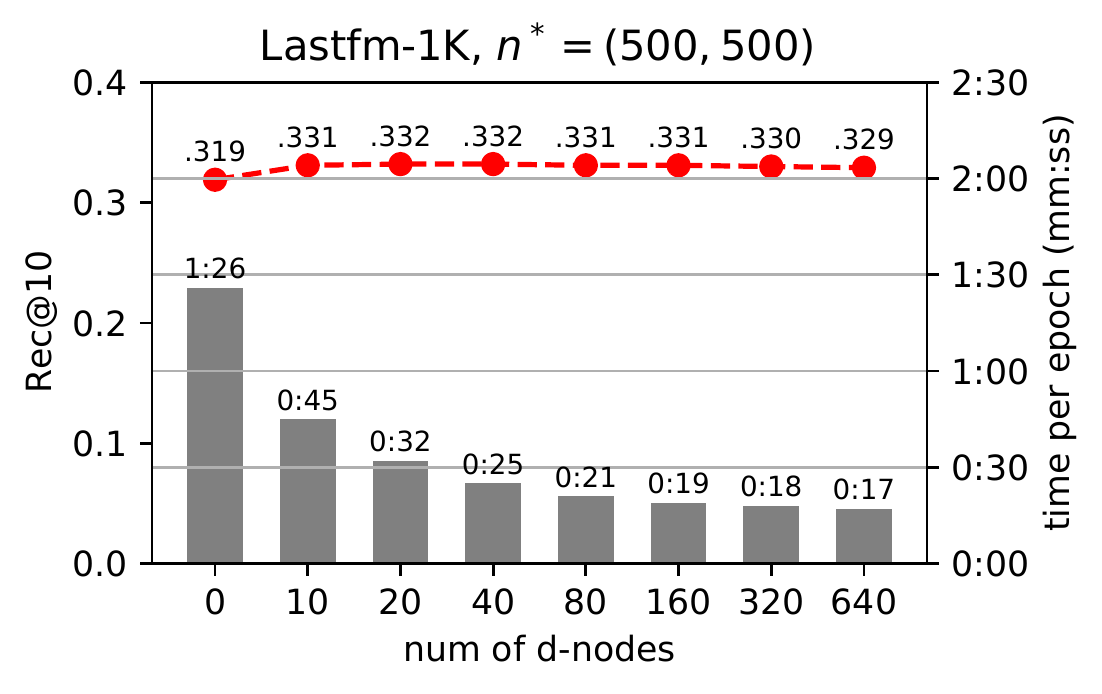}
    &  
    \includegraphics[width=0.3\textwidth]{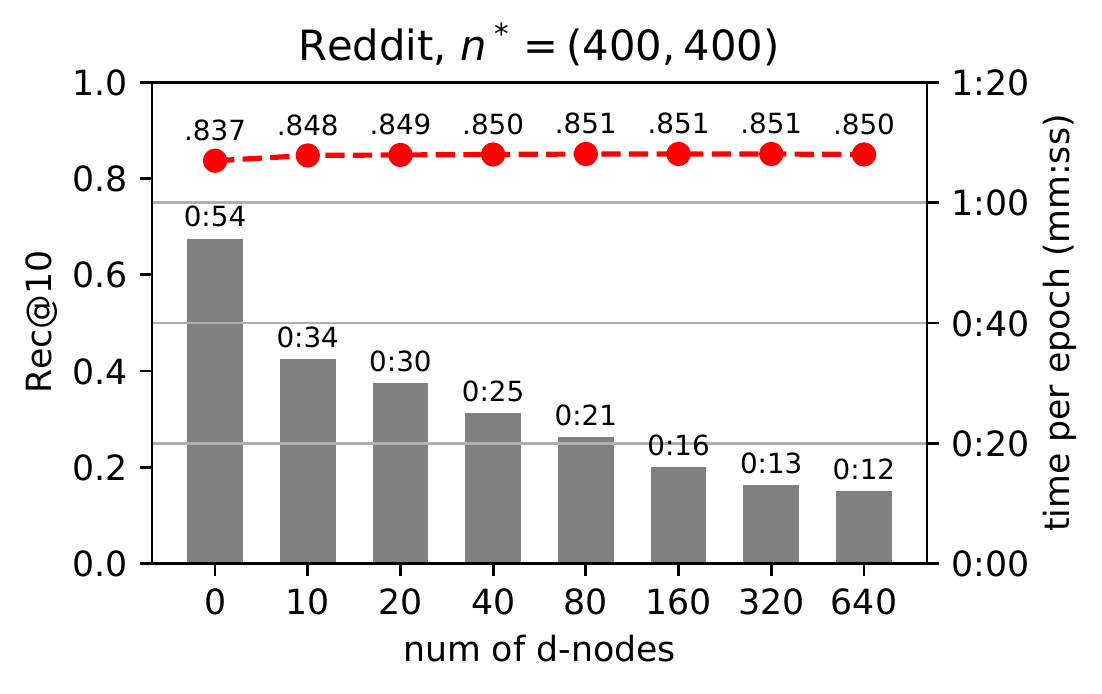}
    &
    \includegraphics[width=0.3\textwidth]{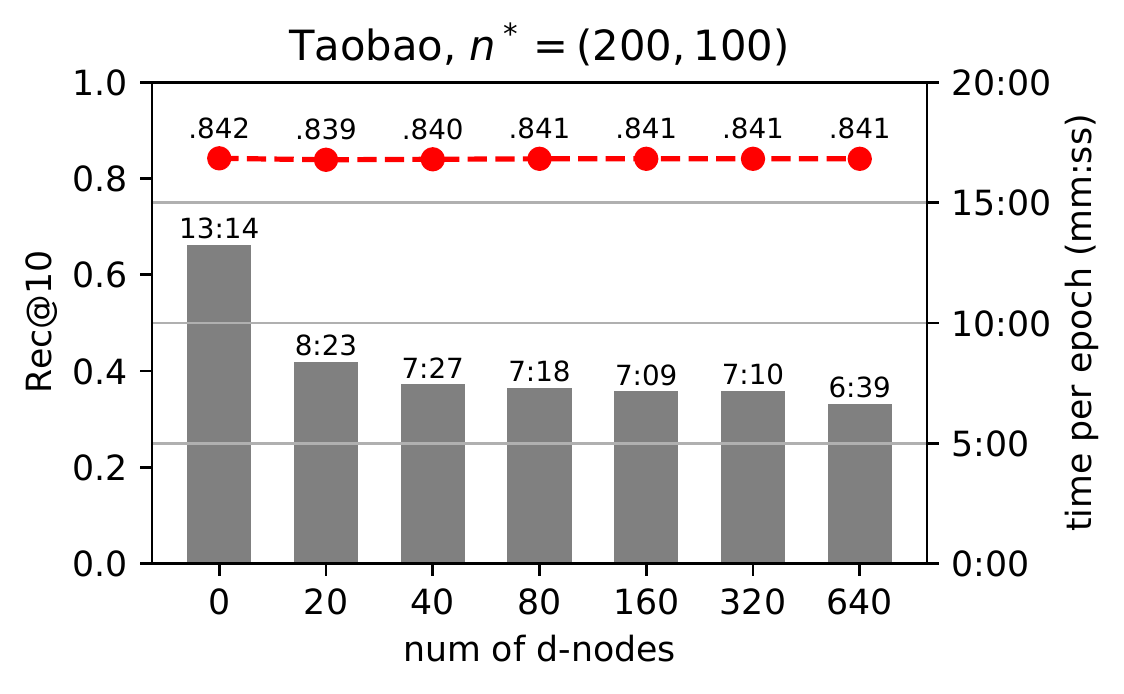}
    \\
    \includegraphics[width=0.3\textwidth]{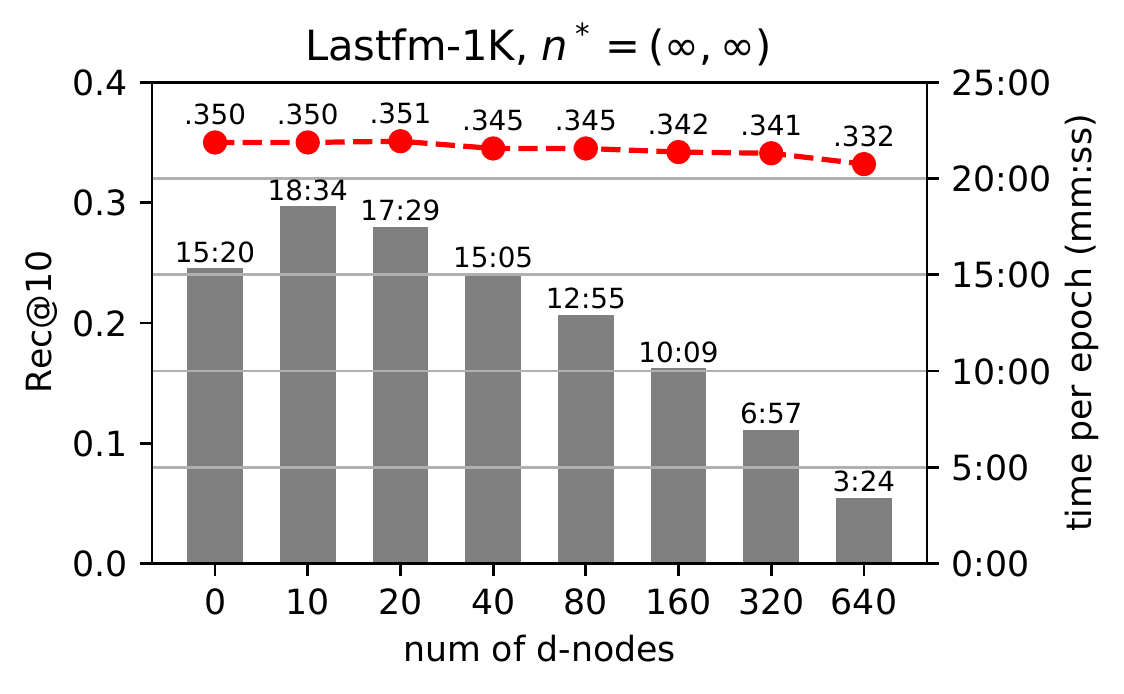}     & 
    \includegraphics[width=0.3\textwidth]{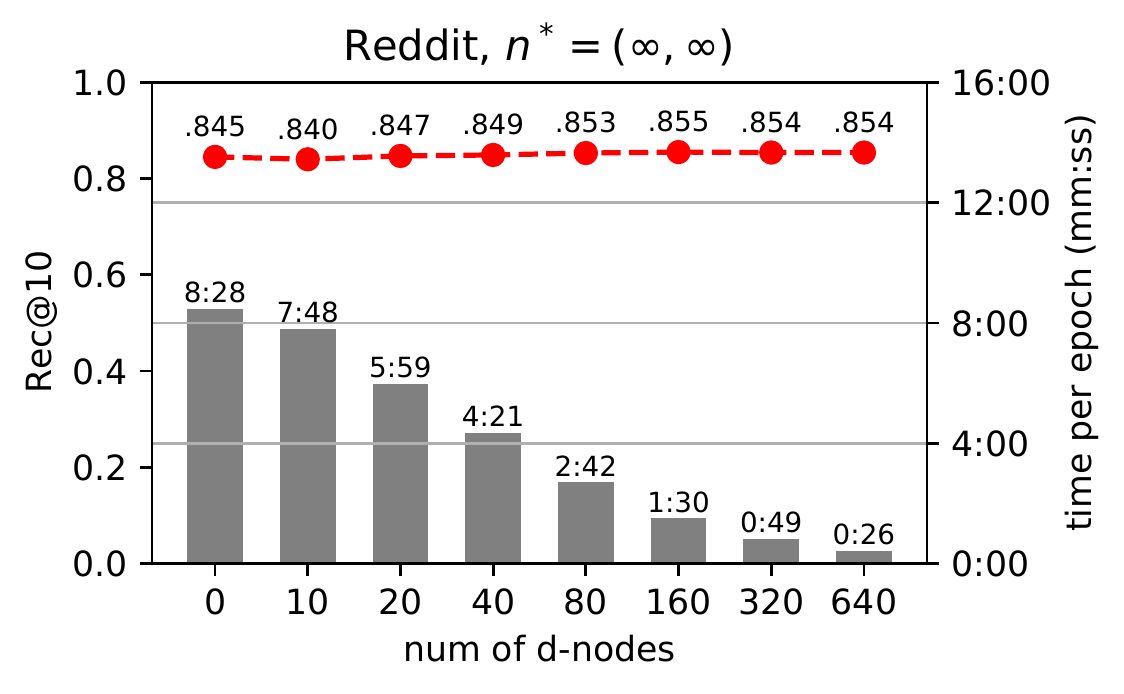}
    &
    \includegraphics[width=0.3\textwidth]{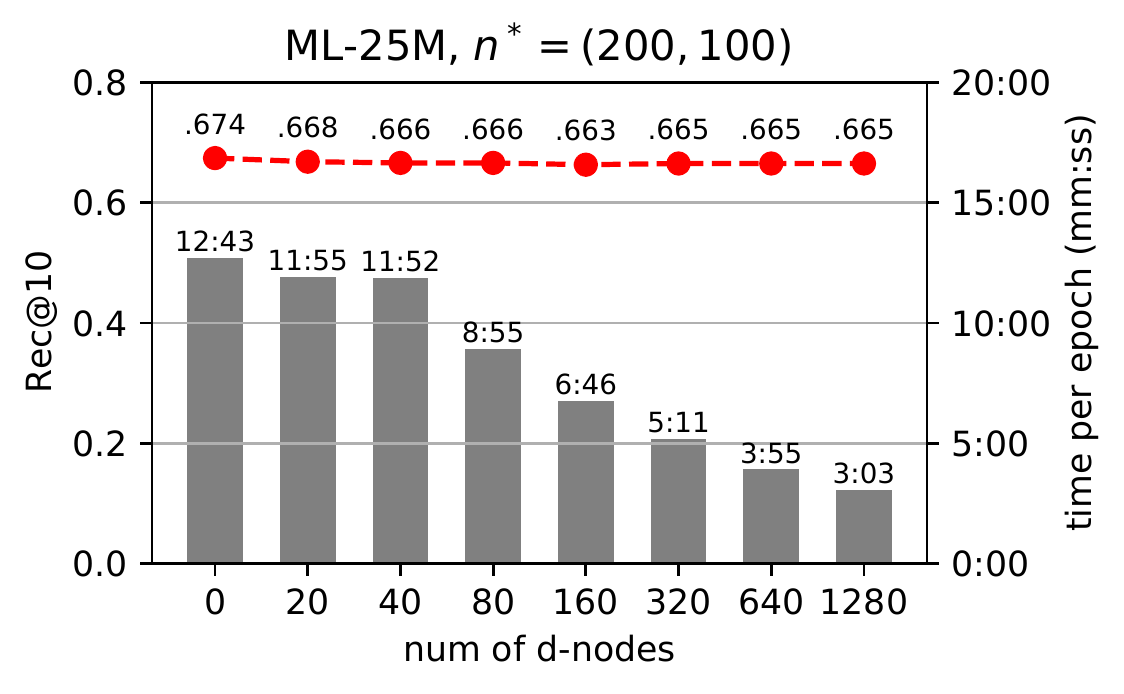}
    \end{tabular}
    \caption{Bar plots indicate the training time (mins:seconds) per epoch. Red scatter points are the Rec@10 performance. Similar figures with MRR performance are given in Appendix~\ref{app:more}.}
    \label{fig:ablation}
\end{figure*}

\textit{(ii) Training efficiency}: In Fig.~\ref{fig:time-mrr},  the $x$-axis is the training time per epoch, which is a measure of training efficiency, and the $y$-axis is the MRR performance on test set. While achieving good performances in MRR, EDGE is very efficient. It has a similar efficiency as the sequence models, and is a lot faster than both GNN-based models and exisiting dynamic graph models, which take 2 hours or more to run on a single epoch.  Furthermore, we observe that EDGE actually takes fewer number of epochs to converge compared to the sequence models. The \#(d-nodes) and active threshold $n^*$ are the same as that in Table~\ref{tab:results}. Results with different configurations will be compared later.

\begin{table}[h]
    \centering
    \caption{Comparison to dynamic graph models.}%
    \label{tab:results-small}
    \begin{tabular}{|@{\hspace{1mm}}c@{\hspace{1mm}}|c@{\hspace{2mm}}c|c@{\hspace{2mm}}c@{\hspace{1mm}}|}
    \hline
    \multirow{2}{*}{method} & \multicolumn{2}{c|}{lastfm-1K} &  \multicolumn{2}{c@{}|}{Reddit}
    \\
    & MRR  & Rec@10   & MRR  & Rec@10 \\
    \hline
    CTDNE & 0.01 & 0.01 & 0.165 & 0.257 \\
    DeepCoevolve & 0.019 & 0.039 & 0.171 & 0.275 \\
    JODIE & {\color{blue}0.195} & {\color{blue}0.307} & {\color{red}0.726} & {\color{blue}0.852} \\
    dynAERNN & 0.021 & 0.038 & 0.157 & 0.301\\
    TGAT & 0.015 & 0.023 & 0.602 & 0.775 \\
    \name & {\color{red}0.199} & {\color{red}0.332} & {\color{blue}0.725} & {\color{red}0.855} 
    \\
    \hline
  \end{tabular} 
\end{table}
\begin{figure}[h]
    \centering
    \includegraphics[width=0.31\textwidth]{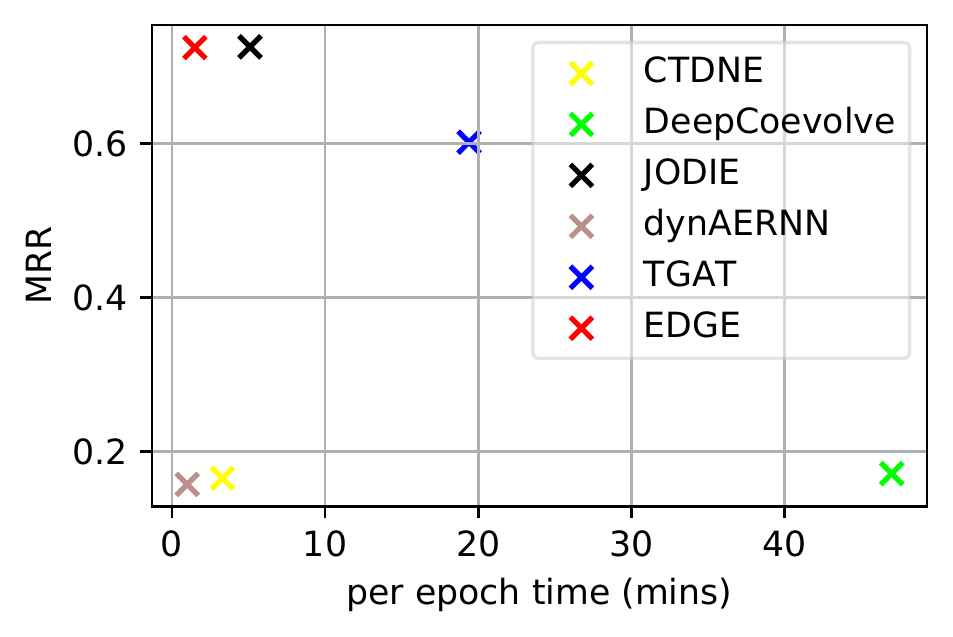}
    \caption{Run-time comparison on Reddit.}
    \label{fig:reddit-time}
\end{figure}

\textbf{Result 2: On small datasets.} Lastfm-1K and Reddit are datasets pre-processed by JODIE which are relatively small so that all dynamic graph models can scale to them. \textit{(i) MRR and Rec@10} are summarized in Table~\ref{tab:results-small}. Overall, EDGE has similar performances as JODIE, which is expected, because the architecture we use in EDGE is similar to that in JODIE. The advantage of EDGE is in efficient training, but on small datasets, JODIE can be well trained, too. \textit{(ii) Runtime:} However, EDGE is more efficient than JODIE.  Fig.~\ref{fig:reddit-time} shows the training time of EDGE when \#(d-nodes) per batch is 160 and the threshold for active nodes is $n^*=\infty$, so all its speed-up  comes from d-nodes.

\textbf{Result 3: Ablation study on the d-nodes.} We validate the effectiveness of EDGE, especially the speed-up from d-nodes, by a series of ablation experiments. Fig.~\ref{fig:ablation} summarizes the results. By gradually increasing the number of d-nodes, we can observe the decrease in training time from the bar plots. The d-nodes are very effective in reducing the computational cost in the sense that they constitute a small portion of the computational nodes, especially on large datasets, and do not have much effect on the prediction performances. We can even observe some improvement in prediction accuracy when using more d-nodes. This might be credited to the easiness of training on shorter sequences.

\begin{table*}[h]
\caption{Longest path comparison on ML-25M.}
    \label{tab:path-length}
    \centering
    \begin{tabular}{|c|c|cccccccc|}
        \hline
         Algorithm~\ref{algo:dnodes} (ours) & number of d-nodes & 0 & 10 & 20 & 40 & 80 & 160 & 320 & 640 \\
         & longest path length & 209.75 & 164.05 & 142.81 & 115.01 & 89.63 & 66.12 & 45.47 & 29.43 \\
         \hline \hline
         cut-by-time & number of parts & 1 & 2& 4& 8& 16& 32 & 64 & 128\\
          &  number of d-nodes & 0 & 134 & 250 & 346 & 429 & 506 & 588 & 701 \\
          & longest path length & 209.75 & 202.6 & 198.95 & 196.29 & 193.02 & 190.84 & 187.39 & 179.51 \\
          \hline
    \end{tabular}
\end{table*}
\textbf{Result 4: Reduction in the path length.} To verify that our selection algorithm has actually reduce the longest path in the computational DAG, we explicitly compute its length after decoupling the selected d-nodes. For comparison, we also implement a comparatively simpler approach, called `cut-by-time'. It selects the d-nodes to divide the DAG into multiple disconnected parts based on the time intervals. Results are reported in Table~\ref{tab:path-length}, which reveals a few advantages of our heuristic. First, we can easily control the number of d-nodes. However, in the case of cut-by-time or other algorithms that cut the DAG into separate parts, the number of d-nodes is determined by the number of parts. Second, our heuristic is a lot more effective, in terms of reducing the length of the longest path.

\textbf{Result 5: Ablation study on the active nodes.} What if active nodes do not use static embeddings? The main purpose of using static embeddings is to improve the efficiency and ease of training, which is especially important for large datasets. To demonstrate this, we gradually increase the node splitting threshold $n^*$ to see the change in run-time and accuracy. Note that when $n^*$ increases, fewer nodes are categorized as active nodes to use static embeddings. For the extreme case when 
$n^*=\infty$, all nodes are treated as inactive and do not use static embedding. It is observed in Table~\ref{tab:ablation-act-taobao} that using more active nodes can effectively improve training efficiency. Especially, on Movielesns-25M, the efficiency will be unacceptable without using static embeddings for active nodes.  In terms of accuracy, on Taobao, the case of $(n_1^*,n_2^*)$=(200,100) has achieved as good performance as other thresholds. On movielens-25M, it seems 
$(n_1^*,n_2^*)$=(400,200) is a better choice for accuracy and efficiency trade-off.
\begin{table}[h]
    \centering
    \caption{Results of using fewer nodes as active nodes.}
    \label{tab:ablation-act-taobao}
    \begin{tabular}{|c|c|c|c|}
    \multicolumn{4}{c}{Taobao} \\
    \hline
    $(n_1^*,n_2^*)$ & time per epoch & MRR & Rec@10\\
    \hline
    (200,100)     &  13.14 mins & 0.675 & 0.642 \\
    (400,200)     & 13.84 mins & 0.674 & 0.841 \\
    (800,400) & 14.52 mins & 0.674 & 0.841 \\
    (1600,800) & 15.57 mins & 0.675 & 0.841 \\
    ($\infty$, $\infty$) & 27.50 mins & 0.674 & 0.841 \\
    \hline
    \multicolumn{4}{c}{ML-25M} \\
    \hline
    $(n_1^*,n_2^*)$ & time per epoch & MRR & Rec@10\\
    \hline
    (200,100)     &  12.72 mins & 0.397 & 0.674 \\
    (400,200)     & 22.79 mins & 0.404 & 0.683 \\
    (800,400) & 1 hr 4 mins & 0.409 & 0.687 \\
    (1600,800) & 3 hr 1 mins & 0.406 & 0.688 \\
    ($\infty$, $\infty$) & 20 hr 54 mins & - & - \\
    \hline
    \end{tabular}
    
\end{table}

\section{Conclusion and discussion}
\label{sec:conclusion}
In this paper, we have proposed an efficient framework, EDGE, to address the computational challenges of learning large-scale dynamic graphs. We evaluated EDGE on large-scale item ranking tasks, and showed that its performances outperform several classes of SOTA methods. The models discussed in this paper are mainly based on node-wise updates by an recurrent operator when new
interactions appear, which lack a GNN-like aggregation from a node’s neighbors. Future works include adaptation of this efficient algorithm to more sophisticated aggregations, and also exploration on effective training algorithm for constrained optimization.

\bibliographystyle{unsrt}  
\bibliography{bibfile}

\newpage
\appendix

\section{Dataset details}
\label{app:dataset}

\subsection{Overall Description}
\textbf{Taobao} dataset contains user-item interaction data from November 25 to December 03, 2017 (one week). 
The details of this dataset can be found at \url{https://tianchi.aliyun.com/dataset/dataDetail?dataId=649}.

\textbf{Movielens-25M} dataset contains about 25 million rating activities from MovieLens, from January 09, 1995 to November 21, 2019. The details of this dataset can be found at \url{https://grouplens.org/datasets/movielens/25m/}.

\textbf{Lastfm-1K} dataset contains music listening data of 1000 users. We use the dataset prepared by \citep{kumar2019predicting}. The details can be found at \url{https://github.com/srijankr/jodie} and \url{http://ocelma.net/MusicRecommendationDataset/lastfm-1K.html}.

\textbf{Reddit} dataset contains one month
of posts made by users on subreddits~\citep{baumgartner2020pushshift}. We use the dataset filtered by \citep{kumar2019predicting}, which contains the 1,000
most active subreddits as items and the 10,000 most active users. The details can be found at \url{https://github.com/srijankr/jodie}.
 
\subsection{Node Degree Distribution }
 
 \begin{figure}[h!]
    \centering
    \includegraphics[width=0.4\textwidth]{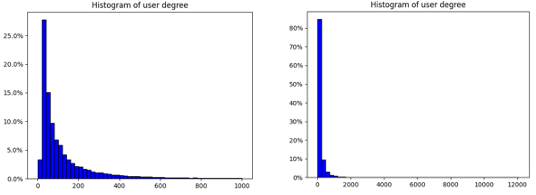}\\
    (a) ML-25M
    \\
    \includegraphics[width=0.4\textwidth]{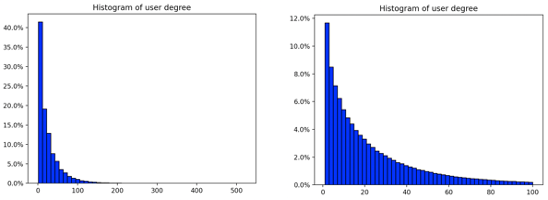}\\
    (b) Taobao
    \caption{Distribution of the number of observed interactions in the training dataset.}
    \label{fig:degree}
\end{figure}

\section{Baseline specifications}
\label{sec:baseline-mod}

Some of the compared baseline methods are not directly scalable to large scale interaction graphs, or originally designed for ranking task. To make the baselines runnable on large and sparse graphs and comparable to our proposed method, we have made a few improvements to the released implementations.
\begin{itemize}[leftmargin=*]
\item \textbf{GRU4Rec \& DIEN \& MIMN}: We changed the data batching from per-user based to  per-example based, to better adapt for time ordered interaction data and to better make the FULL use of the training data. Besides, the implementation of GRU4Rec follows implementation by the authors of MIMN paper, which includes some modifications compared to the original version of GRU4Rec released by their authors, and should be expected to perform better.

\item \textbf{GCMC \& GCMC-SAGE \& GCMC-GAT}: We changed the loss function from Softmax over different ratings to Softmax over [true item, 10 random selected items], to better adapt to the ranking task.

\item \textbf{dynAERNN}: These two methods are originally designed for discrete graph snapshots, while our focused tasks are continues interaction graphs. We manually transformed the interactions graph into 10 snapshots, with equal edge count increments.  

For the downstream ranking task, we followed the evaluation method used in dySat \citep{sankar2020dysat}: after the node embeddings are trained, we train a logistic regression classifier to predict link between a pair of user / item node embedding using Hadmard operator. The logits on test set are used for ranking. For Taobao dataset, we are not able to get results of dynAERNN given the memory constraint.

\item \textbf{JODIE}: we made the following adaptions: 1) replaced the static one-hot representation to 64-dimensional learnable embeddings because the number of nodes are too large that the original implementation will have the out-of-memory issue; 2) add a module to deal with categorical feature via a learnable embedding table since the original implementation is more suitable for continuous feature; 3) used triplet loss with random negative example rather than original MSE loss, which empirically show improvements.

\item \textbf{TGAT}: We notice that the performance of TGAT is particularly bad on Taobao. We spend efforts on checking the experiments to make sure there is no bug. Finally, we identify two issues of using this method on Taobao dataset. Firstly, TGAT is an inductive model, which could potentially give a better generalization ability for \textit{newly observed} nodes in the test set, but on the other hand it is not expressive enough to make use of the large-scale dataset. 
Secondly, the original implementation in TGAT does not deal with categorical feature, but in the Taobao dataset, the only feature is a categorical feature for the items. We need to add a module to map the categorical feature to a learnable feature embedding table (similar to how we modify the feature module in JODIE) to achieve a better performance. With this feature embedding module, we were able to increase its performance on Taobao to about 0.086 (MRR) and 0.164 (Rec@10).
\end{itemize}


\section{Ablation study results}
\label{app:more}

We present the complete results for the ablation study blow.

\begin{figure*}[h!]
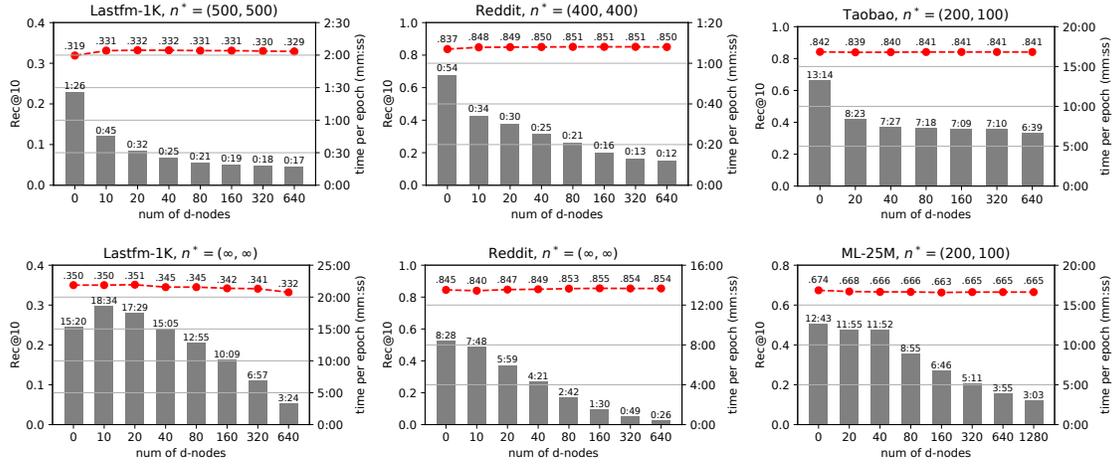

    \centering
    \begin{tabular}{@{}c@{}c@{}c@{}}
    \includegraphics[width=0.3\textwidth]{Figs/lastfm_dttr.pdf}
    &  
    \includegraphics[width=0.3\textwidth]{Figs/reddit_dttr.pdf}
    &
    \includegraphics[width=0.3\textwidth]{Figs/taobao_dttr.pdf}
    \\
    \includegraphics[width=0.3\textwidth]{Figs/lasinf_dttr.pdf}     & 
    \includegraphics[width=0.3\textwidth]{Figs/redinf_dttr.pdf}
    &
    \includegraphics[width=0.3\textwidth]{Figs/mvlens_dttr.pdf}
    \end{tabular}
    \caption{Bar plots indicate the training time (mins:seconds) per epoch. Red scatter points are the \textbf{Rec@10} performance.}
\end{figure*}
\begin{figure*}[h!]
    \centering
    \begin{tabular}{@{}c@{}c@{}c@{}}
    \includegraphics[width=0.3\textwidth]{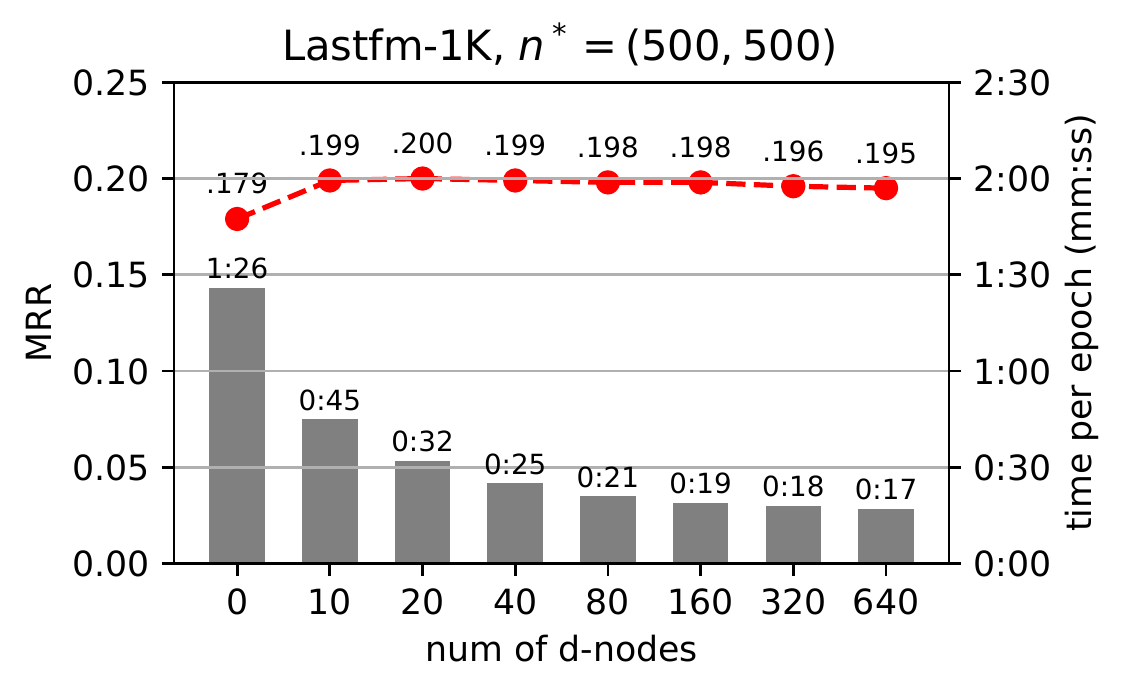}
    &  
    \includegraphics[width=0.3\textwidth]{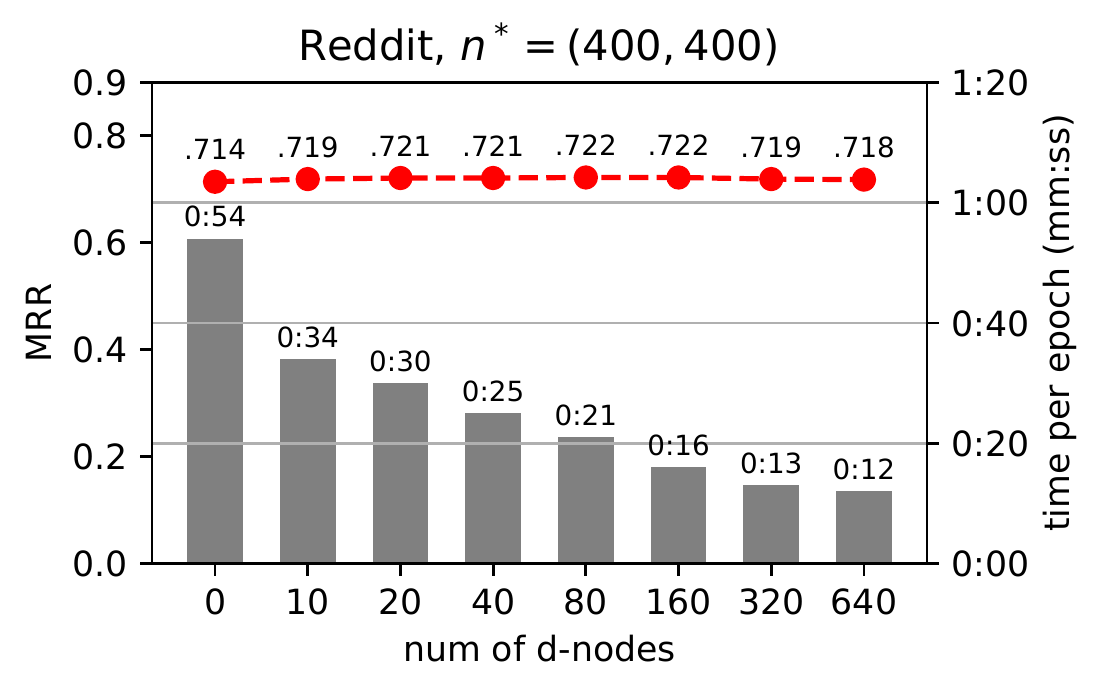}
    &
    \includegraphics[width=0.3\textwidth]{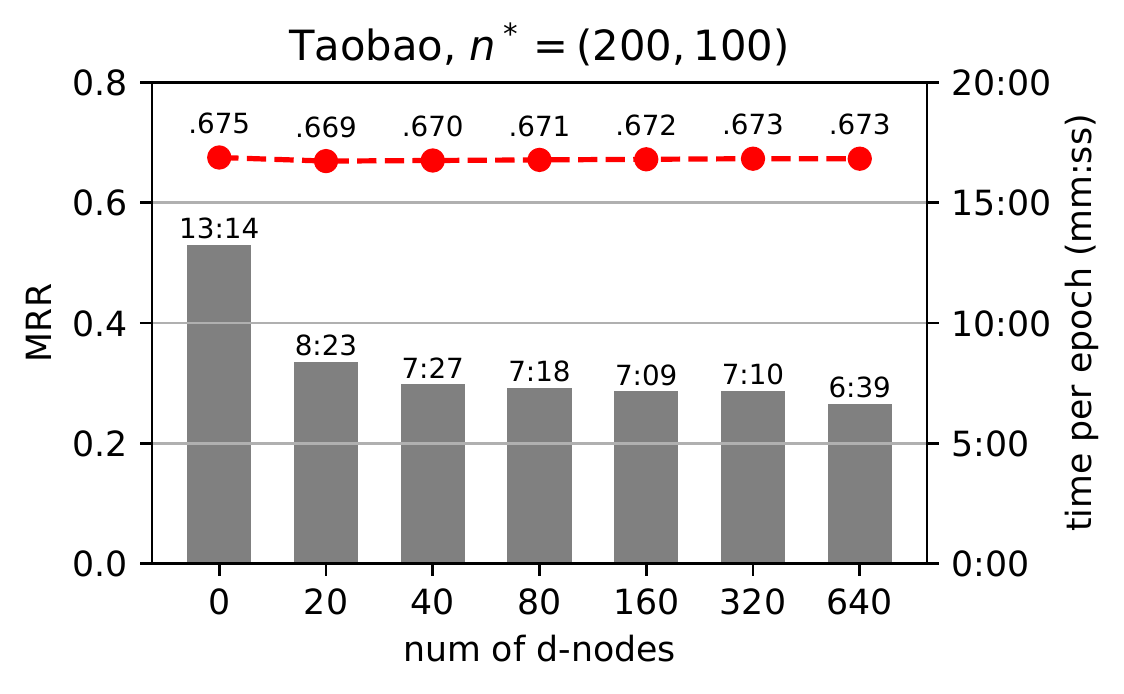}
    \\
    \includegraphics[width=0.3\textwidth]{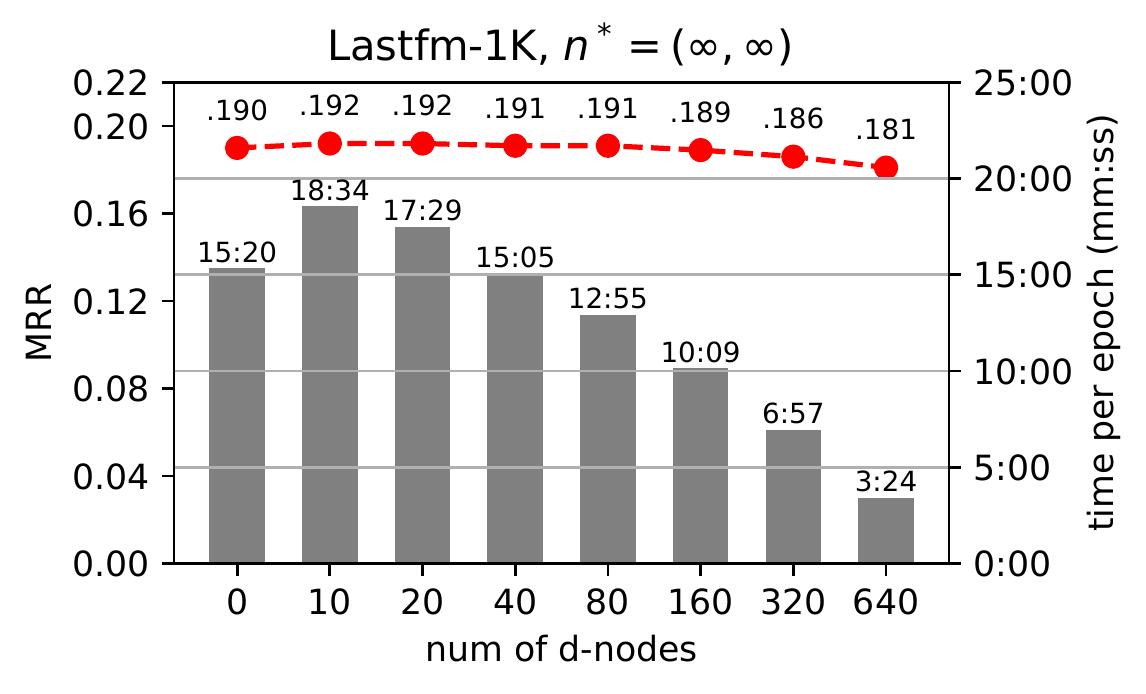}     & 
    \includegraphics[width=0.3\textwidth]{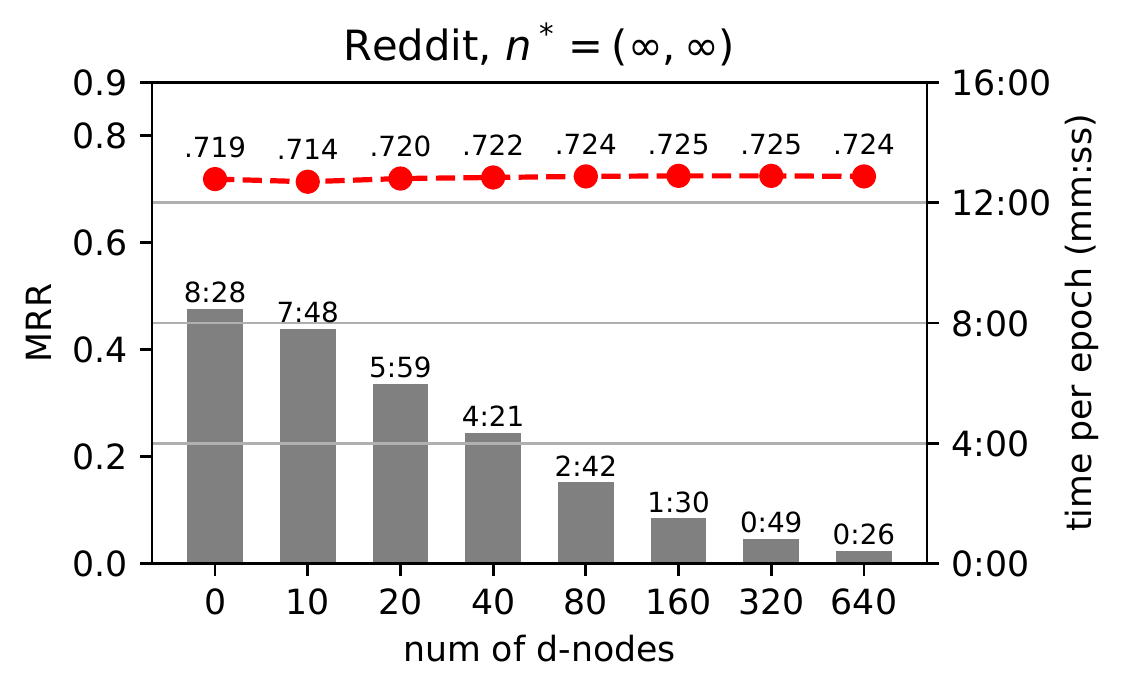}
    &
    \includegraphics[width=0.3\textwidth]{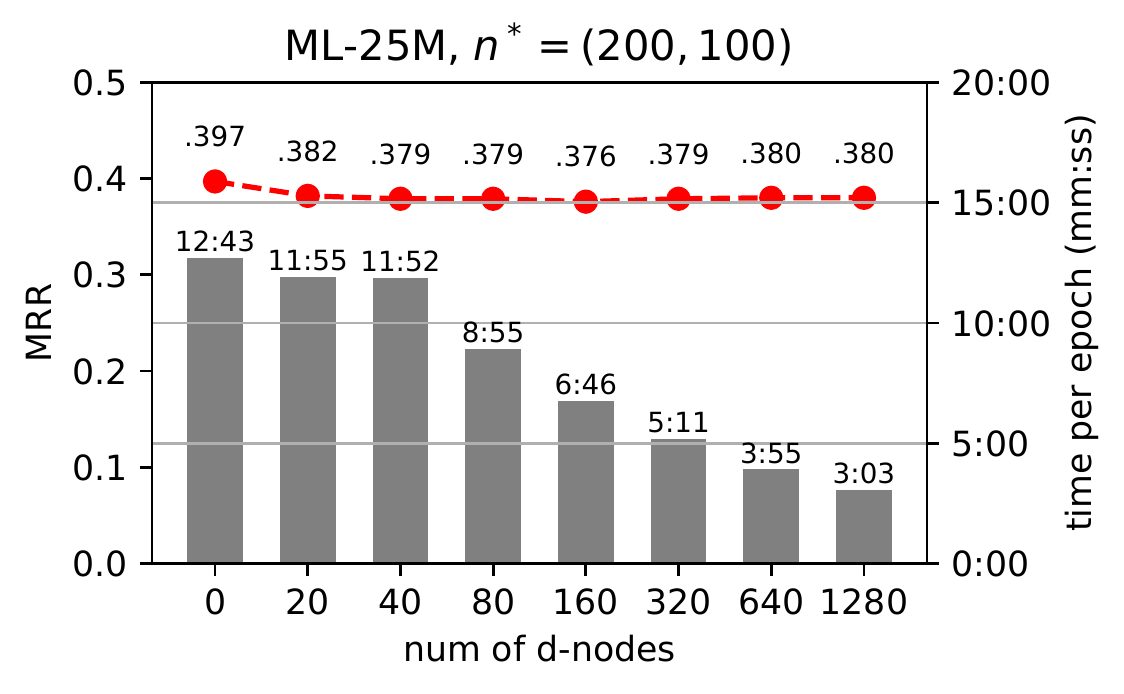}
    \end{tabular}
    \caption{Bar plots indicate the training time (mins:seconds) per epoch. Red scatter points are the \textbf{MRR} performance.}
\end{figure*}

\section{Computing resources}
Experiments on large-scale datasets are run on NVIDIA Quadro RTX 6000 (24 GB memory) or on NVIDIA Tesla v100 (16 GB memory). Experiments on small datasets are either run on NVIDIA GeForce 2080 Ti (11 GB memory) or on NVIDIA Tesla v100 (16 GB memory). Each individual job is run on a single GPU.
\end{document}

%% file: Figs/math_commands.tex
\usepackage{amsmath,amsfonts,bm}

\def\eqref#1{equation~\ref{#1}}

\def\1{\bm{1}}

\def\vphi{{\bm{\phi}}}
\def\vpsi{{\bm{\psi}}}
\def\vxi{{\bm{\xi}}}

\def\vh{{\bm{h}}}

\def\vw{{\bm{w}}}

\DeclareMathAlphabet{\mathsfit}{\encodingdefault}{\sfdefault}{m}{sl}
\SetMathAlphabet{\mathsfit}{bold}{\encodingdefault}{\sfdefault}{bx}{n}

\def\gD{{\mathcal{D}}}
\def\gE{{\mathcal{E}}}
\def\gF{{\mathcal{F}}}

\def\gH{{\mathcal{H}}}

\def\gL{{\mathcal{L}}}

\def\gO{{\mathcal{O}}}

\def\gV{{\mathcal{V}}}

\def\sN{{\mathbb{N}}}

\newcommand{\R}{\mathbb{R}}

\DeclareMathOperator*{\argmax}{arg\,max}

\newcommand{\rbr}[1]{\left(#1\right)}

\def\bb{\begin{equation}} \def\ee{\end{equation}}